\documentclass[conference]{IEEEtran}
\IEEEoverridecommandlockouts
\usepackage{cite}
\usepackage{amsmath,amssymb,amsfonts}
\usepackage{algorithmic}
\usepackage{graphicx}
\usepackage{textcomp}
\usepackage{url}
\def\BibTeX{{\rm B\kern-.05em{\sc i\kern-.025em b}\kern-.08em
    T\kern-.1667em\lower.7ex\hbox{E}\kern-.125emX}}
\begin{document}

\title{The Unlikely Duel: Evaluating Creative Writing in LLMs through a Unique Scenario*
%{\footnotesize %\textsuperscript{*}Note: Sub-titles are not captured in Xplore and
%should not be used}
\thanks{We acknowledge the European Research Council (ERC), which has funded this research under the Horizon Europe research and innovation programme (SALSA, grant agreement No 101100615); Grant SCANNER-UDC (PID2020-113230RB-C21) funded by MICIU/AEI/10.13039/501100011033; Xunta de Galicia (ED431C 2020/11); and Centro de Investigación de Galicia ‘‘CITIC’’, funded by the Xunta de Galicia through the collaboration agreement between the Consellería de Cultura, Educación, Formación Profesional e Universidades and the Galician universities for the reinforcement of the research centres of the Galician University System (CIGUS).}
}

\author{\IEEEauthorblockN{Carlos Gómez-Rodríguez}
\IEEEauthorblockA{\textit{Universidade da Coruña, CITIC} \\
\textit{Department of CS and IT}\\
A Coruña, Spain \\
carlos.gomez@udc.es}
\and
\IEEEauthorblockN{Paul Williams}
\IEEEauthorblockA{\textit{School of Business \& Creative Industries} \\
\textit{University of the Sunshine Coast}\\
Sunshine Coast, Australia \\
pwillia3@usc.edu.au}
}

\maketitle

\begin{abstract}

This is a summary of the paper ``A Confederacy of Models: a Comprehensive Evaluation of LLMs on Creative Writing'', which was published in Findings of EMNLP 2023.
We evaluate a range of recent state-of-the-art, instruction-tuned large language models (LLMs) on an English creative writing task, and compare them to human writers. 
For this purpose, we use a specifically-tailored prompt (based on an epic combat between Ignatius J. Reilly, main character of John Kennedy Toole's ``A Confederacy of Dunces'', and a pterodactyl) to minimize the risk of training data leakage and force the models to be creative rather than reusing existing stories. The same prompt is presented to LLMs and human writers, and evaluation is performed by humans using a detailed rubric including various aspects like fluency, style, originality or humor. Results show that some state-of-the-art commercial LLMs match or slightly outperform our human writers in most of the evaluated dimensions. Open-source LLMs lag behind. Humans keep a close lead in originality, and only the top three LLMs can handle humor at human-like levels.
\end{abstract}


\section{Note}

This is a summary of the paper ``A Confederacy of Models: a Comprehensive Evaluation of LLMs on Creative Writing'', which was published in Findings of EMNLP 2023~\cite{GomWilEMNLP2023}.

\section{Introduction}

In the last few years, large language models (LLMs) have shown remarkable performance at many language processing and generation tasks~\cite{zhao2023survey}. This has motivated research on evaluating their performance at various tasks, as well as comparing it to humans.

In the work summarized here~\cite{GomWilEMNLP2023}, we provided a comprehensive evaluation of LLMs on a creative writing task, which was lacking. Our methodology featured two key points to guarantee reliability (in relative terms, considering the subjectivity of the task) and avoid the common pitfalls of LLM benchmarking: (i) our evaluation was based entirely on human judgments, using a detailed creative writing rubric; and (ii) our setting was purposefully designed to prevent training data contamination, a pervasive problem in LLM evaluation, especially when working with closed models~\cite{sainz-etal-2023-nlp}. For this purpose, we used a custom-made zero-shot prompt with a bizarre scenario:

\begin{quote}
Write an epic narration of a single combat between Ignatius J. Reilly and a pterodactyl, in the style of John Kennedy Toole.
\end{quote}

The prompt combines a character who is well-defined, but appeared in a single book (``A Confederacy of Dunces'') and does not seem to appear in fan fiction; with a very different challenge from any scene in the book. Thus, copying or heavily drawing from existing stories in the training data should not be helpful to complete the task to a competent level, so we avoid training data contamination while forcing the models to be creative.\footnote{Note, however, that a drawback of this approach is that the prompt cannot reliably be reused in the future, since after publication of the paper the prompt and the corpus of generated stories are online and could be used for training LLMs. To keep ensuring data leakage in future studies with the same methodology, one would need to create a fresh prompt.} Our prompt has the extra advantages of being challenging, requesting a distinct literary style, and featuring humor, which is especially challenging for LLMs~\cite{jentzsch-kersting-2023-chatgpt}.

\section{Materials and methods}

\paragraph{Models} We compared every instruction-tuned LLM that we could find by the study's cutoff date of April 20, 2023. This selection yielded 12 LLMs: Alpaca, Bard, Bing Chat, ChatGPT with GPT 3.5, ChatGPT with GPT 4, Claude, Dolly 2.0, GPT4All-J, Koala, OpenAssistant, StableLM and Vicuna. More details (specific model versions and sizes, and detailed selection criteria) are provided in~\cite{GomWilEMNLP2023}.

\paragraph{Evaluation rubric} We used a rubric based on previous work by one of the authors of this paper~\cite{carey2022re}, but adapted to the specific task. The rubric evaluates ten distinct aspects in a 1-10 scale: readability, key narrative elements, structural elements, plot logic, creativity, incorporation of John Kennedy Toole style, epic genre, description and credibility of combat, main characters, and dark humor.

\paragraph{Human writers} Were Honours and postgraduate Creative Writing students that volunteered for the task. %As in any evaluation comparing humans and LLMs, care must be taken not to assume that our sample is representative of human performance as a whole.

\paragraph{Evaluation methodology} We prompted each LLM 5 times, from a fresh state, with the prompt above. This allows us to measure reliability and control for the inherent randomness in LLM output. We then asked 5 human writers to write their own stories following the same prompt, and with a similar length to the LLM-generated stories. 10 raters performed the human evaluation. Each rater was given 13 stories (one by a human and one by each of the LLMs) but they did not know which was which.

\begin{figure}[tbp]
    \centering
    \includegraphics[width=1.0\linewidth]{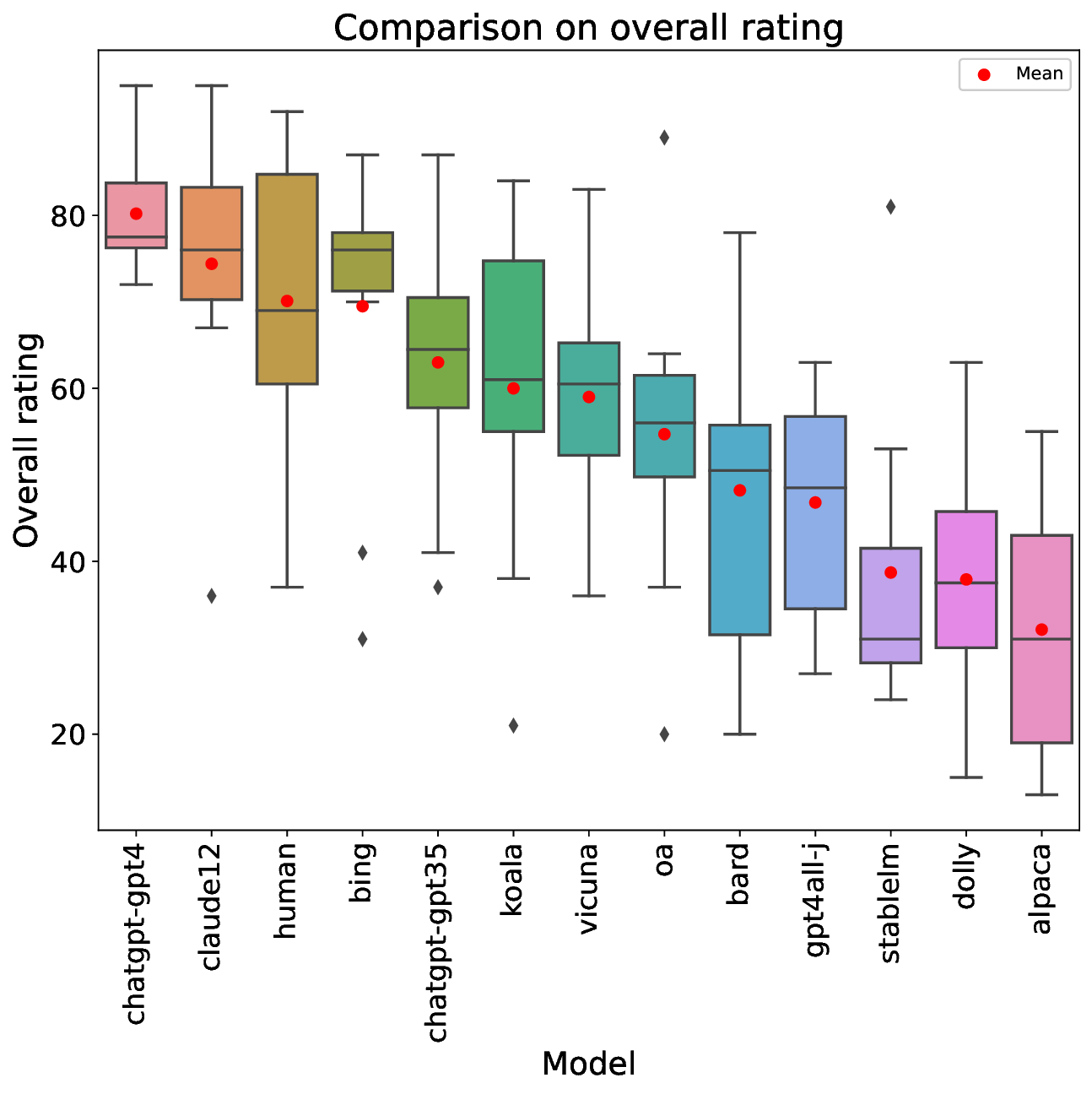}
    \caption{Comparison of overall ratings for stories %generated 
    by humans and 12 LLMs, sorted left to right by mean overall rating. Boxes show median, quartiles Q1-Q3, and whiskers at 1.5 IQR, with values outside that range plotted as outliers. Filled red circles denote means.}
    \label{fig:box_overall}
\end{figure}

\section{Results}

\paragraph{Agreement} Weighted Cohen's kappa was 0.48 $[0.43, 0.54]$, i.e. moderate agreement, a good result considering the high subjectivity of the task.

\paragraph{General overview} Results for each model (as well as humans) in terms of overall scores (average across all rubric items) are shown in Figure~\ref{fig:box_overall}. ChatGPT with GPT-4 is the overall best model, outrating our human writers and showing remarkable consistency. A significance test on overall rating does not show significant differences between humans and the top 6 LLMs (with the bottom 6 being significantly worse), but a more fine-grained test considering individual rubric items as datapoints does identify GPT-4 as significantly better than our human writers, Claude and Bing as not significantly different, and the remaining LLMs as significantly worse. Finally, it is worth remarking that commercial LLMs have the best results in this analysis, with open-source models lagging clearly behind.

\paragraph{Individual dimensions} We summarize here some key highlights from the results in~\cite{GomWilEMNLP2023}, which also provides plots for all rubric dimensions.

Humor is the most difficult rubric item, with an average score across models of 3.4. There is a clear divide between a few LLMs that can handle humor (GPT-4, Claude and Bing); which score comparably to humans, and the rest, which are far behind in this dimension.

The item measuring creativity and originality is the only one where human writers outperformed all LLMs in our analysis, although the differences between humans and the top three LLMs are not detected as significant in a t-test.

LLMs seem to be especially good at the more technical aspects, like readability and structure, as well as in understanding and habitation of the epic genre, where eight models are rated better (two of them significantly better) than human writers.

\section{Discussion}

Our thorough evaluation of LLMs and comparison to human writers showed  state-of-the-art LLMs slightly outperforming humans at our writing task. While this should not be generalized to claims like ``superhuman storytelling'' (both due to sample size and our writers not necessarily being representative of human writing ability as a whole), it does strongly suggest that the best LLMs can, at the very least, perform the task at a very competent level. Other interesting conclusions are that open LLMs clearly lag behind commercial models; and that LLMs clearly excel at technical aspects whereas humans keep an edge in originality. The corpus with all the stories is available at~\url{https://doi.org/10.5281/zenodo.8435671}.

\section*{Acknowledgment}

We thank our volunteers who participated in the writing and grading of stories, whose names are acknowledged in~\cite{GomWilEMNLP2023} and the corpus above.

\bibliographystyle{plain}
\bibliography{biblio}

\end{document}